\documentclass[sigconf]{acmart}

\usepackage{graphicx}
\usepackage{amsmath}
\usepackage{booktabs}
\usepackage{cancel}

\usepackage{multirow}
\usepackage{xspace}
\usepackage{bm}
\usepackage{wrapfig}
\usepackage{enumitem}
\usepackage{microtype}
\usepackage{subfig}
\usepackage[switch]{lineno}

\def\abbrmodelname{ATOM}
\def\fullmodelname{ATomic mOtion Modeling}

\makeatletter
\DeclareRobustCommand\onedot{\futurelet\@let@token\@onedot}
\def\@onedot{\ifx\@let@token.\else.\null\fi\xspace}

\def\eg{\emph{e.g}\onedot} 
\def\ie{\emph{i.e}\onedot}

\def\etal{\emph{et al}\onedot}
\makeatother

\usepackage[capitalize]{cleveref}
\crefname{section}{Sec.}{Secs.}
\Crefname{section}{Section}{Sections}
\Crefname{table}{Table}{Tables}
\crefname{table}{Tab.}{Tabs.}

\AtBeginDocument{%
  }

\copyrightyear{2023}
\acmYear{2023}
\setcopyright{acmlicensed}\acmConference[MM '23]{Proceedings of the 31st ACM International Conference on Multimedia}{October 29-November 3, 2023}{Ottawa, ON, Canada}
\acmBooktitle{Proceedings of the 31st ACM International Conference on Multimedia (MM '23), October 29-November 3, 2023, Ottawa, ON, Canada}
\acmPrice{15.00}
\acmDOI{10.1145/3581783.3612289}
\acmISBN{979-8-4007-0108-5/23/10}

\begin{document}

\title{Language-guided Human Motion Synthesis with Atomic Actions}

\author{Yuanhao Zhai}
\affiliation{%
  \institution{University at Buffalo}
  \city{Buffalo}
  \state{NY}
  \country{USA}}
\email{yzhai6@buffalo.edu}

\author{Mingzhen Huang}
\affiliation{%
  \institution{University at Buffalo}
  \city{Buffalo}
  \state{NY}
  \country{USA}}
\email{mhuang33@buffalo.edu}

\author{Tianyu Luan}
\affiliation{%
  \institution{University at Buffalo}
  \city{Buffalo}
  \state{NY}
  \country{USA}}
\email{tianyulu@buffalo.edu}

\author{Lu Dong}
\affiliation{%
  \institution{University at Buffalo}
  \city{Buffalo}
  \state{NY}
  \country{USA}}
\email{ludong@buffalo.edu}

\author{Ifeoma Nwogu}
\affiliation{%
  \institution{University at Buffalo}
  \city{Buffalo}
  \state{NY}
  \country{USA}}
\email{inwogu@buffalo.edu}

\author{Siwei Lyu}
\affiliation{%
  \institution{University at Buffalo}
  \city{Buffalo}
  \state{NY}
  \country{USA}}
\email{siweilyu@buffalo.edu}

\author{David Doermann}
\affiliation{%
  \institution{University at Buffalo}
  \city{Buffalo}
  \state{NY}
  \country{USA}}
\email{doermann@buffalo.edu}

\author{Junsong Yuan}
\affiliation{%
  \institution{University at Buffalo}
  \city{Buffalo}
  \state{NY}
  \country{USA}}
\email{jsyuan@buffalo.edu}

\renewcommand{\shortauthors}{Yuanhao Zhai et al.}

\begin{abstract}
  Language-guided human motion synthesis has been a challenging task due to the
  inherent complexity and diversity of human behaviors.
  Previous methods face limitations in generalization to novel actions, often
  resulting in unrealistic or incoherent motion sequences.
  In this paper, we propose \abbrmodelname{} (\fullmodelname{}) to mitigate this
  problem, by decomposing actions into atomic actions, and employing a curriculum
  learning strategy to learn atomic action composition.
  First, we disentangle complex human motions into a set of atomic actions during
  learning, and then assemble novel actions using the learned atomic actions,
  which offers better adaptability to new actions.
  Moreover, we introduce a curriculum learning training strategy that leverages
  masked motion modeling with a gradual increase in the mask ratio, and thus 
  facilitates atomic action assembly.
  This approach mitigates the overfitting problem commonly encountered in previous
  methods while enforcing the model to learn better motion representations.
  We demonstrate the effectiveness of \abbrmodelname{} through extensive
  experiments, including text-to-motion and action-to-motion synthesis tasks.
  We further illustrate its superiority in synthesizing plausible and coherent
  text-guided human motion sequences.
\end{abstract}


%



\maketitle

\begin{figure}[t]
    \includegraphics[width=\linewidth]{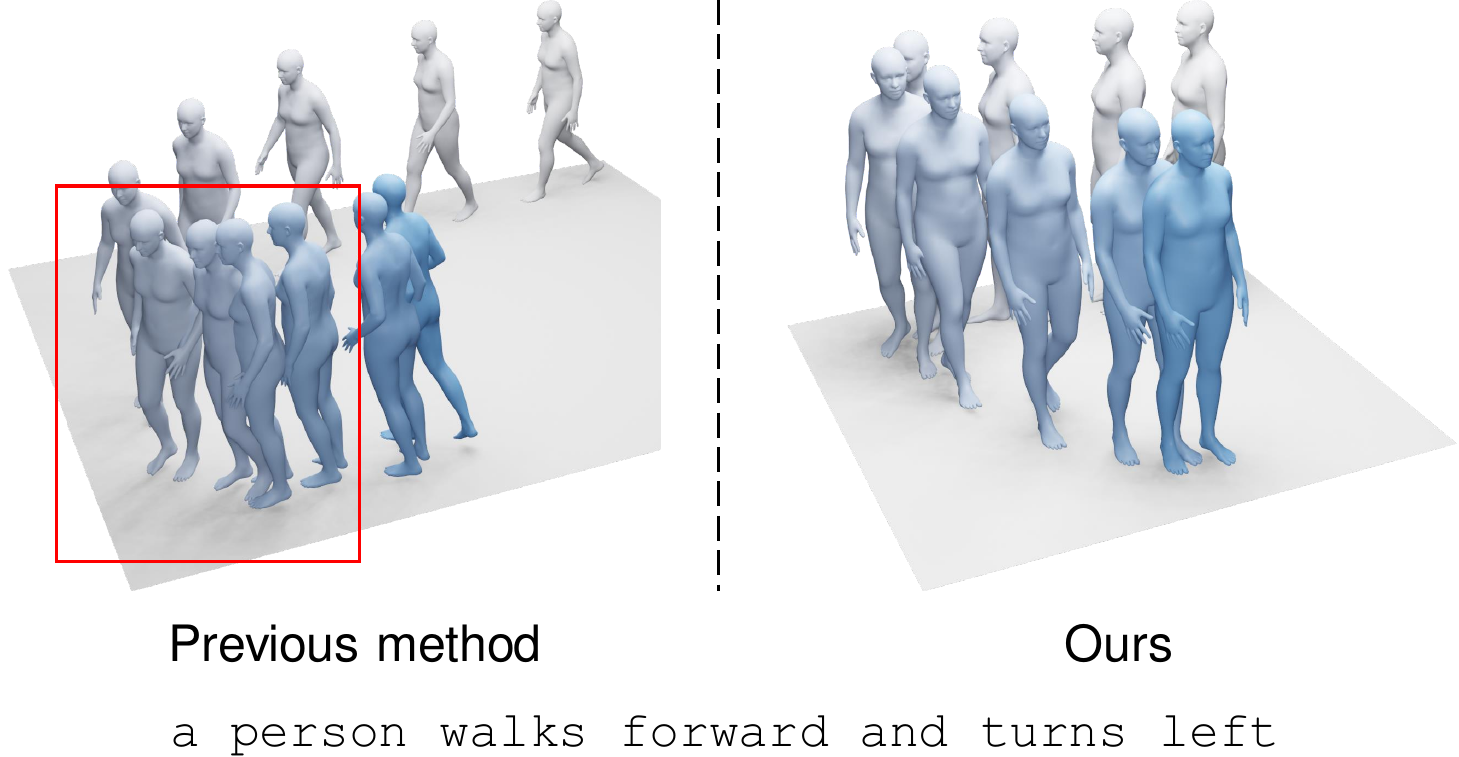}
    \caption{Motion synthesis results comparison. Previous
    T2M~\cite{guo2022generating} generates an unrealistic motion transition between
    ``\texttt{walk forward}'' and ``\texttt{turn left}'' (see motion inside the
    red box), while our \abbrmodelname{} generates coherent motion. The color saturation increases as the motion progresses.}
    \label{fig:teaser}
\end{figure}

\section{Introduction}

Language-guided human motion synthesis is a critical and challenging task, with
widespread applications in virtual reality, video games, animation, and
human-computer interaction. 
The ability to generate realistic and diverse human motions based on textual
descriptions can enable more intuitive control over virtual characters, as well
as seamless integration of user-generated content in various multimedia systems.
Despite previous conditional action generation
models~\cite{zhang2020perpetual,petrovich2021action,du2015hierarchical,yan2018spatial,yu2020structure,kim2020motion,cai2021unified,xie2021physics,dittadi2021full,wang2021synthesizing}
have exploited leveraging closed-set action labels to synthesize human motion,
few of them can work beyond the training action labels, not to mention open
language descriptions.

In recent years, various methods have been proposed for the language-guided
human motion synthesis
task~\cite{ghosh2021synthesis,goutsu2021linguistic,delmas2022posescript,zhang2022motiondiffuse,guo2022generating,athanasiou2022teach,petrovich2022temos,guo2022tm2t,lucas2022posegpt,wang2022humanise,hong2022avatarclip,tevet2022motionclip,tevet2022human}.
While these methods have demonstrated considerable progress in generating human
motions, they may suffer from synthesizing discontinuities and unrealistic
motion transitions when dealing with actions that are not well-represented in
the dataset.
As these methods rely heavily on the availability and diversity of training
data, they may struggle to generate plausible motion sequences for rare or
unseen actions, leading to abrupt transitions and incoherent movement patterns.
Furthermore, when synthesizing complex sequential motion behaviors, these
methods often struggle with capturing dependencies in motion sequences, which is
crucial for ensuring smooth and natural motion transitions between different
types of actions, as illustrated in~\cref{fig:teaser}.

We propose \abbrmodelname{} (\fullmodelname{}), a novel approach for
language-guided human motion synthesis that effectively addresses the
limitations of previous methods.
First, \abbrmodelname{} decomposes actions into atomic components, enabling the
generation of diverse and coherent motion sequences by assembling the learned atomic actions.
Additionally, by employing a masked motion modeling curriculum learning
strategy, our method learns more expressive motion representations and
effectively captures long-range dependencies in motion sequences.

\begin{figure}[t]
    \includegraphics[width=\linewidth]{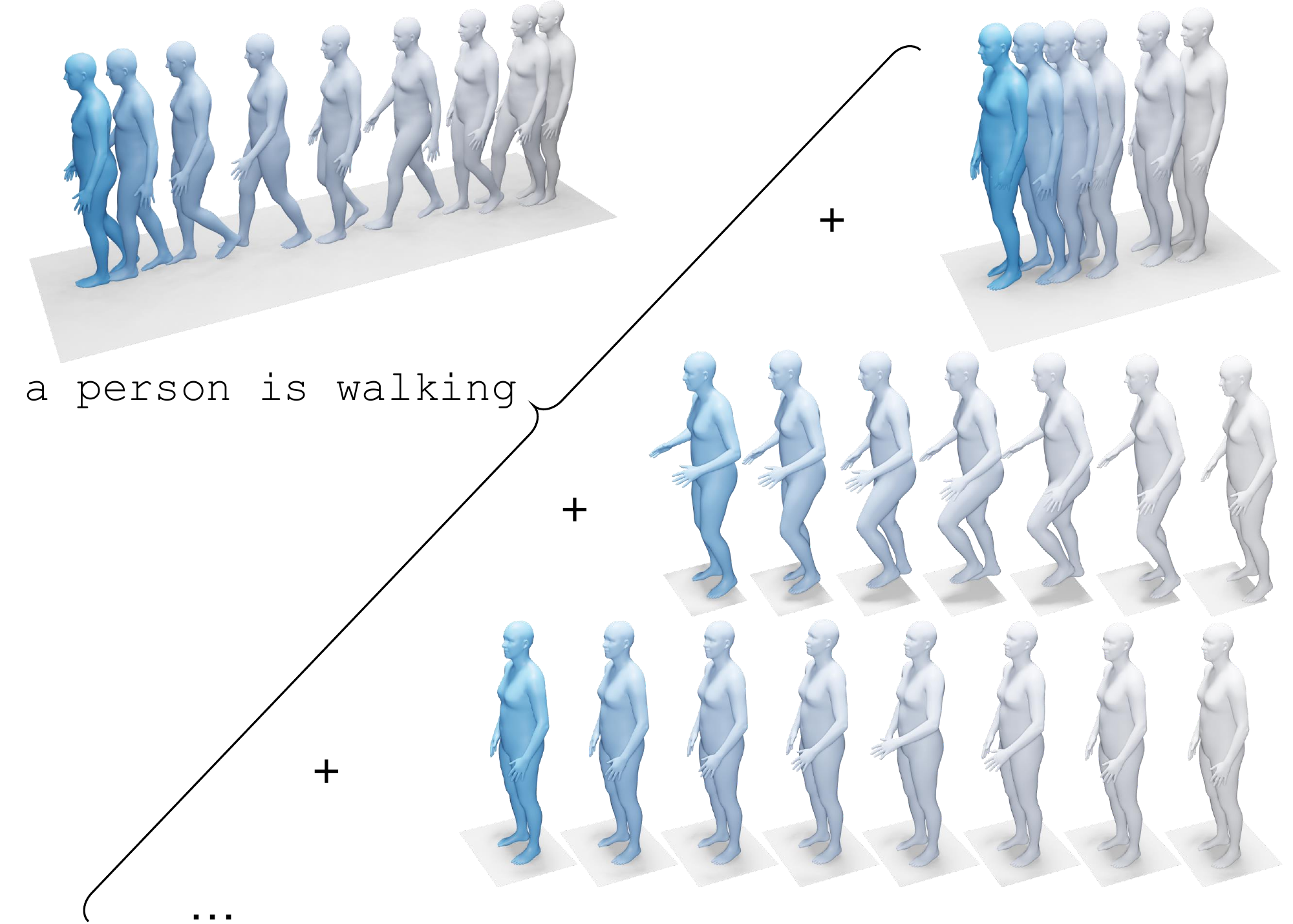}
    \caption{Illustration of atomic action. The action ``\texttt{a person is
    walking}'' can be decomposed into a set of atomic actions: from top to
    bottom, whole body translation, lifting leg, hand movements and so on.}
    \label{fig:illustration}
\end{figure}

In our proposed \abbrmodelname{}, we utilize a transformer-based conditional
variational autoencoder (CVAE) framework~\cite{vaswani2017attention} to achieve
atomic action decomposition and assembly, which provides a more expressive and
flexible representation of human motions.
The atomic action codebook, designed as a set of learnable feature vectors,
serves as the key and value for the cross-attention module in the Transformer
decoder.
These atomic actions, learned in an end-to-end manner, can effectively represent
a wide range of short-term basic human movements, such as raising hands and
lifting legs.
\cref{fig:illustration} showcases an example, where the ``\texttt{walking}''
action can be decomposed into a set of atomic actions.
To ensure the atomicity and effectiveness of the learned actions, we apply two
additional constraints: diversity and sparsity selection constraints.
The diversity constraint encourages the atomic actions to be distinct from each
other, allowing for a richer and more versatile action representation.
The sparsity selection constraint enforces atomic actions to be compact and
focused during composition, ensuring that the learned atomic actions retain
their atomicity.
This constraint aids in the efficient reconstruction of complex actions by
combining a minimal set of atomic actions while preserving their individual
characteristics. 
The combination of the Transformer-based architecture and these constraints
allows our \abbrmodelname{} to efficiently decompose and assemble complex
actions for improved human motion synthesis, ultimately generating more diverse,
coherent and realistic motion sequences based on language input.

Alongside the atomic action decomposition, we incorporate a curriculum learning
strategy into our \abbrmodelname{} to further improve the robustness of
generating coherent and diverse motion sequences while capturing long-range dependencies.
This strategy is based on masked motion modeling and involves gradually
increasing the mask ratio during the training process.
Such a progressive learning approach provides the model with an opportunity to
learn simpler patterns and dependencies in the early stages of training, while
incrementally introducing more complex and challenging aspects of the motion
sequences as training progresses. 
This approach not only mitigates potential convergence issue, but also helps the
model build a more robust and generalizable understanding of motion data, encompassing
both local and long-range motion patterns.
As a result, \abbrmodelname{} is better equipped to synthesize realistic and
diverse human motions based on language inputs, even when dealing with rare or
unseen action labels/descriptions.

To summarize, our contributions are as follows:
\begin{itemize}[leftmargin=*,noitemsep,topsep=0pt]
    \item We introduce a transformer-based CVAE framework that decomposes
    complex actions into a set of atomic actions, enabling more effective
    representation and manipulation of motion sequences. This approach allows
    for generating diverse and coherent motion sequences, even for rare
    or unseen action labels/descriptions.
    \item We incorporate a curriculum learning strategy based on masked motion
    modeling, which gradually increases the mask ratio during training. This
    strategy enables \abbrmodelname{} to better capture dependencies in motion
    sequences, ensuring smooth and natural motion transitions between different
    actions.
    \item We provide a comprehensive evaluation of \abbrmodelname{},
    demonstrating its effectiveness in generating coherent and diverse motion
    sequences. Our method significantly improves over previous approaches
    in the text-to-motion and action-to-motion tasks.
\end{itemize}

\section{Related Work}

\noindent \textbf{Human motion synthesis} \quad
Directly synthesizing human motion has always been a challenging yet ideal task
in computer vision and graphics.
As the field has progressed, methods have evolved from generating skeleton-based
motion
synthesis~\cite{du2015hierarchical,yan2018spatial,yu2020structure,kim2020motion,cai2021unified}
to more realistic SMPL motion
synthesis~\cite{petrovich2021action,xie2021physics, dittadi2021full,
wang2021synthesizing}, bringing us closer to generating authentic human motions
in real-world scenarios~\cite{zhang2021learning,luan2021pc,luan2023high}.
Tilmanne~\etal{}~\cite{tilmanne2010expressive} employed Gaussian distributions
to model the variability of walk cycles for each emotion and the length of each
cycle.
Zhang \etal{}\cite{zhang2020perpetual} further explored generating unbounded
human motion through a cross-conditional, two-stream variational RNN
architecture.
CSGN~\cite{yan2019convolutional} jointly modeled structures in temporal and
spatial dimensions, allowing bidirectional transforms between the latent and
observed spaces to handle semantic manipulation of action sequences.
SA-GCNs~\cite{yu2020structure} proposed a variant of GCNs that leverages the
self-attention mechanism to adaptively sparsify a complete action graph in the
temporal space.
ACTOR~\cite{petrovich2021action} learned an action-aware latent representation
for human motions by training a transformer-based VAE.
ActFormer~\cite{petrovich2021action} focused on multi-person interactive
actions, combining the solid spatio-temporal representation capacity of the
Transformer, the generative modeling superiority of GANs, and the inherent
temporal correlations from latent prior.
INR~\cite{cervantes2022implicit} employs variational implicit neural
representations to generate variable-length sequences.
Recently, a growing body of work has emerged, focusing on leveraging natural
language to synthesize human
motions~\cite{ahuja2019language2pose,ghosh2021synthesis,goutsu2021linguistic,delmas2022posescript,zhang2022motiondiffuse,guo2022generating,athanasiou2022teach,petrovich2022temos,guo2022tm2t,lucas2022posegpt,wang2022humanise,hong2022avatarclip,tevet2022motionclip,tevet2022human}.
Specifically, Language2Pose~\cite{ahuja2019language2pose} learns a joint
embedding of language and pose decoder to generate pose sequences.
Text2gesture~\cite{bhattacharya2021text2gestures} exploits relevant
biomechanical features for body expression to create emotive body gestures.
Hier~\cite{ghosh2021synthesis} introduces a self-supervised method for
generating long-range behaviors.
T2M~\cite{guo2022generating} employs a curated language encoder to learn crucial
words and a duration estimator to synthesize human motions of varying durations.

\noindent \textbf{Language-guided generation} \quad
Language-guided generation establishes a connection between visual
representation and semantic space, enabling more precise control and increased
creative possibilities.
Synthesizing motion from language, especially when dealing with multiple, varied
actions is a challenging task compared to generating images with specific
action labels.
Numerous prior language-guided generation methods have focused on image
generation~\cite{zhang2017stackgan,li2020unicoder,nam2018text,liu2020describe,xia2021tedigan,li2020manigan}.
Reed~\etal{}~\cite{reed2016generative} proposed using a generative adversarial
network~\cite{goodfellow2014generative} (GAN) conditioned on text embeddings for
image synthesis.
DALL·E~\cite{ramesh2021zero} employed a discrete variational autoencoder (dVAE)
to generate diverse images based on text embeddings from
GPT-3~\cite{brown2020language}.
The recently proposed CLIP~\cite{radford2021learning} jointly learns a
multi-modal vision-language embedding space with impressive capabilities.
Leveraging the power of CLIP~\cite{radford2021learning},
StyleCLIP~\cite{patashnik2021styleclip} extends StyleGAN~\cite{karras2019style}
to a language-driven generation model using their proposed CLIP-guided mapper.
Meanwhile, HairCLIP~\cite{wei2021hairclip} generates manipulated images from
given CLIP~\cite{radford2021learning} text embeddings, further demonstrating the
potential of language-driven generation.

\section{Method}

\begin{figure*}[t]
    \includegraphics[width=0.85\linewidth]{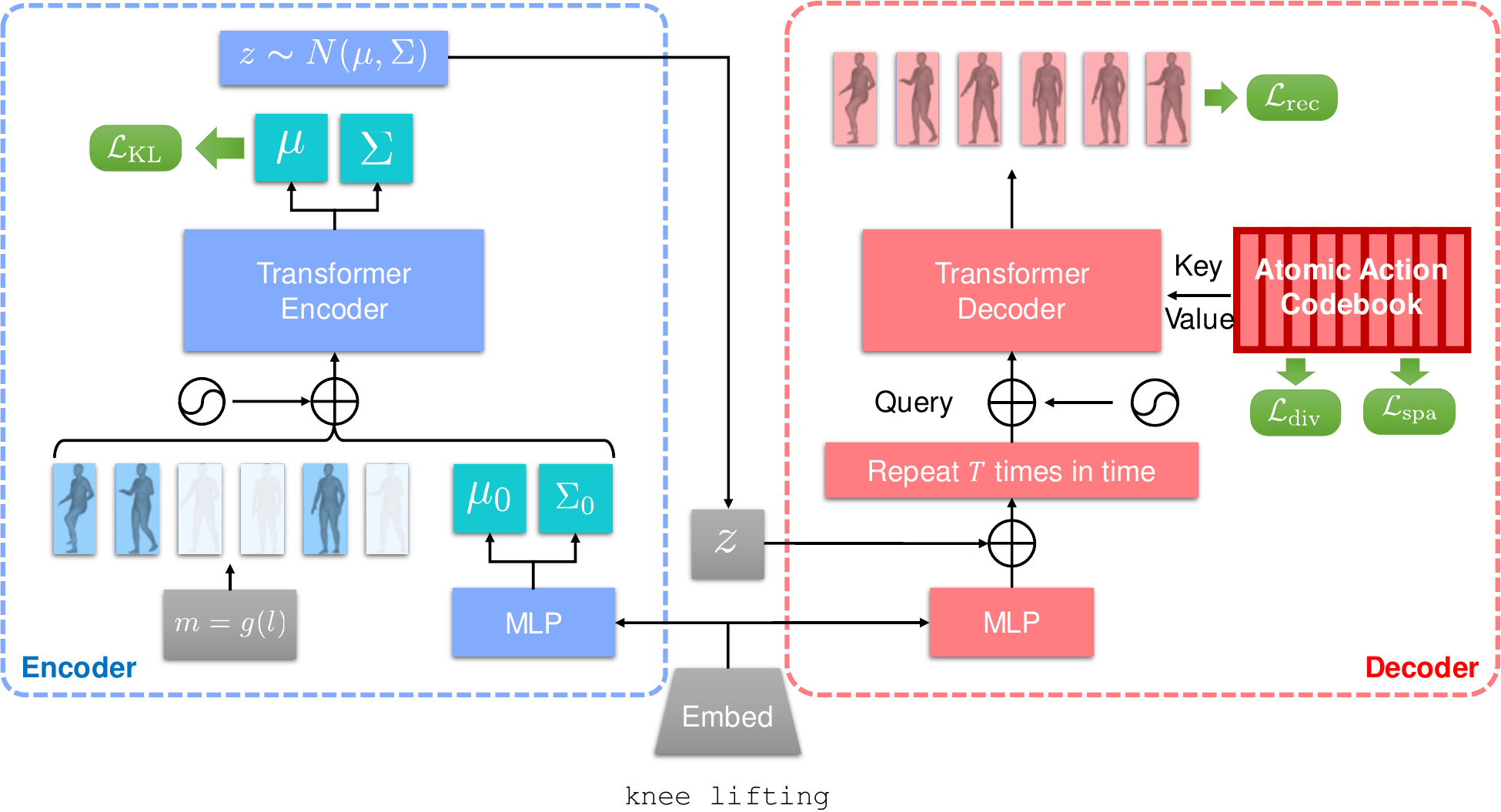}
    \caption{Framework overview. Our \abbrmodelname{} is composed of an encoder
    and a decoder. The encoder processes text embedding of the action label
    and masked motion sequence, outputting a latent vector $z$. The decoder
    receives this latent vector $z$ along with the text embedding. A set of
    learnable atomic actions, referred to as the atomic action codebook, is fed
    into the decoder as key and value for the cross-attention layers. As a
    result, the generated motion sequences are assembled from the atomic
    actions.}
    \label{fig:framework}
\end{figure*}

\subsection{Problem Formulation}

The task of language-guided human motion synthesis aims to generate motion
sequences that accurately represent the given textual description.
During the training phase, the inputs consist of label-motion pairs $\{(y_i,
\bm{M}_i)\}$, where the label can be either a natural language description
(text-to-motion) or a discrete action class (action-to-motion).
The motion representation $\bm{M}_i = [\bm{p}_1, \dots, \bm{p}_T]$ is a sequence
of human body representations with length $T$, where $\bm{p}_t$ denotes the
human body representation at time $t$.
The human body representation can adopt various forms, such as 3D joint
locations or SMPL parameters~\cite{loper2015smpl}.
During the testing phase, the objective is to synthesize motion sequences based
on the input language description or action class.

\subsection{Conditional Transformer VAE}
We utilize a CVAE-based framework, as shown in~\cref{fig:framework}, which aligns motion representations with categorical conditions.
Our \abbrmodelname{} consists of an encoder and a decoder, implemented using Transformer encoder and decoder~\cite{vaswani2017attention}.
The encoder captures the input motion sequence's underlying structure, transforming it into a compact latent representation.
The decoder uses the latent representation and text embedding to generate a realistic human motion sequence corresponding to the given condition.

\noindent \textbf{Encoder} \quad
The encoder accepts the conditional embedding of the label $y$ and the motion
representation sequence $[\bm{p}_t]$, and computes the Gaussian distribution
parameters $\mu$ and $\Sigma$ for the motion latent space.
A latent variable $\bm{z}$ is then sampled from $N(\mu, \Sigma)$ using the
reparameterization trick~\cite{kingma2013auto}.
We prepend two tokens, $\mu_0$ and $\Sigma_0$, to the input, allowing their corresponding outputs to be regarded as the Gaussian distribution parameters.
These two input tokens, $\mu_0$ and $\Sigma_0$, are derived from the input
embedding through a three-layer multilayer perceptron (MLP).
To ensure compatibility, all motion representations $\bm{p}_{t}$ are transformed
to the same dimension as $\mu_0$ and $\Sigma_0$ using a linear layer before
entering the Transformer encoder.

\noindent \textbf{Decoder} \quad
Given a latent vector $\bm{z}$, we first add a conditional bias to it to
incorporate the categorical information.
This bias is action-specific and learned from the input embedding through a
three-layer MLP.
Subsequently, the sum is repeated $T$ times in the temporal dimension, and
sinusoidal positional encoding~\cite{vaswani2017attention} is added as the query
input to the Transformer decoder.
A set of learnable atomic actions is provided to the Transformer decoder as key
and value inputs, enabling the query to be reconstructed using the atomic
actions.
The output of the decoder is the reconstructed motion representation
$\hat{\bm{M}} = [\hat{\bm{p}}_t]$, where $\hat{\bm{p}}_t$ the predicted body representation at time $t$.
It is worth noting that the decoder generates the entire motion sequence in one
shot, as opposed to the autoregressive approach used in previous
works~\cite{petrovich2021action}.

\noindent \textbf{Learning Objectives} \quad
The learning objectives of the CVAE consist of two components: a reconstruction
loss $\mathcal{L}_\text{rec}$ and a Kullback-Leibler (KL) divergence loss
$\mathcal{L}_{\text{KL}}$.
The reconstruction loss is employed to minimize the discrepancy between the
original motion representation $\bm{M}$ and the reconstructed motion
representation $\hat{\bm{M}}$ through a mean square error loss:
\begin{equation}
    \mathcal{L}_{\text{rec}} = \frac{1}{T}\sum_{t=1}^{T} \left \| \bm{p}_t -
    \hat{\bm{p}}_t \right \|_2^2.
\end{equation}
On the other hand, the KL divergence loss $\mathcal{L}_{\text{KL}}$ minimizes
the distribution difference between the estimated posterior $N(\mu, \Sigma)$ and
the prior normal distribution $N(0, I)$.
Thus, the CVAE training loss $\mathcal{L}_{\text{CVAE}}$ is a weighted sum of
the two terms:
\begin{equation}
    \mathcal{L}_{\text{CVAE}} = \mathcal{L}_{\text{rec}} + w_{\text{KL}}
    \mathcal{L}_{\text{KL}},
\end{equation}
where $w_{\text{KL}}$ is a weighting hyperparameter.

\subsection{Atomic Action Codebook}

The motivation for using atomic actions in our method stems from the observation
that human motions, despite their apparent complexity, can often be decomposed
into more specific, repetitive, and atomic elements.
By breaking down complex actions into a series of atomic actions, our model can
more effectively learn the underlying structure of human motion and capture the
relationships between different actions.
Furthermore, this decomposition facilitates the generation of diverse and
realistic motion sequences, as well as the synthesis of novel actions by
recombining the learned atomic elements.
It also enables our method to better generalize across different action classes
and leverage the power of textual descriptions in guiding the synthesis process.

Building on the idea of atomic actions, we introduce a learnable atomic action
codebook, which serves as a basis for representing and generating complex human
motions within our Transformer-based architecture.
This codebook consists of a collection of atomic actions that can be combined
and assembled in various ways to produce a diverse range of motion sequences.
Formally, the codebook is implemented as a learnable matrix $\bm{A} \in
\mathbb{R}^{N\times D}$, where $N$ represents the number of atomic actions, and
$D$ denotes the hidden dimension.
Each row of the matrix corresponds to an atomic action, capturing its unique
characteristics in the latent space.
The codebook is integrated into the Transformer decoder as key and value,
enabling the model to selectively attend to relevant atomic actions during the
decoding process.
This design allows the input conditional embedding to be efficiently
reconstructed by combining the most appropriate atomic actions based on the
cross-attention mechanism in the decoder, leading to more accurate and diverse
motion synthesis. Our experiments in \cref{tab:atomic-action} and \cref{fig:ablation-qua} show the effectiveness of our codebook design compared to the original CVAE.

\begin{table*}[h!]
\centering
\caption{Quantitative results comparison on the HumanML3D test set. $\rightarrow$
indicates results are better if they are closer to the real motion, and $\pm$ indicates 95\% confidence interval.}
\label{tab:humanml3d}
\resizebox{0.675\linewidth}{!}{%
\begin{tabular}{c|ccccc}
    \hline 
    Method & FID $\downarrow$ & Diversity $\rightarrow$ & MultiModality $\uparrow$ & R Precision (top3) $\uparrow$ & MultiModal Dist $\downarrow$\tabularnewline
    \hline 
    Real Motion & $0.002^{\pm .000}$ & $9.503^{\pm .065}$ & - & $0.797^{\pm .002}$ & $2.974^{\pm .008}$\tabularnewline
    \hline 
    Language2Pose~\cite{ahuja2019language2pose} & $11.02^{\pm .046}$ & $7.676^{\pm .058}$ & - & $0.486^{\pm .002}$ & $5.296^{\pm .008}$\tabularnewline
    Text2Gesture~\cite{bhattacharya2021text2gestures} & $7.664^{\pm .030}$ & $6.409^{\pm .071}$ & - & $0.345^{\pm .002}$ & $6.030^{\pm .008}$\tabularnewline
    Hier~\cite{ghosh2021synthesis} & $6.532^{\pm .024}$ & $8.332^{\pm .042}$ & - & $0.552^{\pm .004}$ & $5.012^{\pm .018}$\tabularnewline
    T2M~\cite{guo2022generating} & $\bm{0.455^{\pm .003}}$ & $9.175^{\pm .002}$ & $2.219^{\pm.074}$ & $\bm{0.736^{\pm .002}}$ & $\bm{3.347^{\pm .074}}$\tabularnewline
    MoCoGAN~\cite{tulyakov2018mocogan} & $94.41^{\pm .021}$ & $0.462^{\pm .008}$ & $0.019^{\pm .000}$ & $0.106^{\pm .001}$ & $9.643^{\pm .006}$\tabularnewline
    Dance2Music~\cite{lee2019dancing} & $66.98^{\pm .016}$ & $0.725^{\pm .011}$ & $0.043^{\pm .001}$ & $0.097^{\pm .001}$ & $8.116^{\pm .006}$\tabularnewline
    Ours & $1.691^{\pm .031}$ & \bm{$9.312^{\pm .011}}$ & $\bm{2.884^{\pm .130}}$ & $0.569^{\pm .004}$ & $5.970^{\pm .004}$\tabularnewline
    \hline 
\end{tabular}%
}
\end{table*}

\noindent \textbf{Learning Objectives} \quad
To ensure the effectiveness of the atomic action codebook in generating a wide
range of motion sequences, we introduce two objectives for its learning:
diversity constraint $\mathcal{L}_{\text{div}}$ and sparsity constraint
$\mathcal{L}_{\text{spa}}$.
The diversity constraint ensures that the learned atomic actions are diverse and
unique, so that the learned atomic actions are diverse enough to represent
different actions.
By promoting diversity in the codebook, our model can generate more
realistic and rich motion sequences that cover a broad spectrum of human
actions.
Formally, the diversity constraint is formulated as follows:
\begin{equation}
    \mathcal{L}_{\text{div}} = \| \bm{A} \bm{A}^{\top} - \bm{I} \|_F,
\end{equation}
where $\bm{A}$ is the atomic action codebook, $\bm{I}$ is the identity
matrix, and $\|\cdot\|_F$ is the Frobenius norm.
This objective encourages the learned atomic actions to be orthogonal, promoting
diversity and preventing codebook redundancy.
Our experiments in column ``diversity" of \cref{tab:humanml3d}, \cref{tab:kit},
\cref{tab:uestc}, and \cref{tab:humanact12} show that the diversity of our
method is higher than the previous approaches and closer to real motions on
multiple datasets.

The sparsity constraint promotes the use of a sparse set of atomic actions to
represent complex motions, enhancing the atomicity and robustness of the learned
atomic actions.
This ensures that generated motion sequences are concise and
meaningful while maintaining the interpretability and generalizability of the
atomic action codebook.
We enforce the sparsity constraint by maximizing the maximal attention values
in each cross-attention layer:
\begin{equation}
    \mathcal{L}_{\text{spa}} = -\sum_{l}\sum_{h}\max (\bm{H}_{l,h}),
\end{equation}
where $\bm{H}_{l,h}$ is the attention map of the cross-attention for the $h$-th
head in layer $l$.
This loss encourages the model to focus on a few dominant
atomic actions when reconstructing motion sequences, leading to a sparser and
more interpretable atomic action codebook.

By incorporating the atomic action constraints, the total loss
$\mathcal{L}_{\text{total}}$ of our \abbrmodelname{} is as follows:
\begin{equation}
    \mathcal{L}_{\text{total}} = \mathcal{L}_{\text{CVAE}} + w_{\text{div}}
    \mathcal{L}_{\text{div}} + w_{\text{spa}} \mathcal{L}_{\text{spa}},
\end{equation}
where $w_{\text{div}}$ and $w_{\text{spa}}$ are weighting hyperparameters.

\subsection{Masked Motion Modeling Curriculum Learning}
Drawing inspiration from the effective representation learning of masked image
autoencoders~\cite{he2022masked}, we introduce masked motion modeling, a
technique that involves temporally masking a random portion of the input motion
sequence at a ratio $r$, and subsequently requiring the model to reconstruct the
entire motion sequence.
Masked motion modeling is essential in the motion synthesis task because it
encourages the model to learn robust, context-aware motion representations by
forcing it to fill in the missing information.
This approach results in a more robust and generalized understanding of the
underlying motion structure, ultimately leading to more realistic motion
synthesis.

To further enhance the learning process, we incorporate curriculum learning by
progressively increasing the mask ratio as the training progresses according to
a growth function $g(l) \in \{ r, rl/L, r(l/L)^2, re^{l/L - 1} \}$, where $l$ is
the current training epoch and $L$ is the total number of epochs.
As a result, in the beginning, a lower mask ratio allows the model to learn
basic motion patterns and capture fundamental structures in the motion data.
As the mask ratio increases, the model is exposed to more challenging and
complex motion sequences, promoting its ability to infer missing information and
makes better use of the learned atomic actions.
This curriculum strategy enables a more effective and stable learning
experience, ultimately improving motion synthesis performance.

\section{Experiments}
\label{sec:experiments}

We evaluate our method in three different settings: \emph{text-to-motion},
\emph{action-to-motion}, and \emph{zero-shot action-to-motion}.
Text-to-motiona and zero-shot action-to-motion are to generate human motion
given an input text prompt; while action-to-motion generates motion given an
input action class in the form of one-hot label.

\noindent \textbf{Dataset} \quad
Five datasets are used for the experiments.
For the text-to-motion evaluation, we use the HumanML3D~\cite{guo2022generating}
and KIT~\cite{plappert2016kit} datasets.
The HumanML3D dataset~\cite{guo2022generating}, a recent development, is
generated through the reannotation of AMASS~\cite{mahmood2019amass} and
HumanAct12~\cite{guo2020action2motion} datasets.
It consists of $14,616$ motions accompanied by $44,970$ textual descriptions.
The KIT dataset~\cite{plappert2016kit} consists of $3,911$ motions and the
corresponding descriptions.

For the action-to-motion evaluation, we use two different datasets:
HumanAct12~\cite{guo2020action2motion}, UESTC~\cite{ji2018large}, and
NTU94~\cite{shahroudy2016ntu,liu2019ntu}.
The UESTC dataset~\cite{ji2018large} comprises $25$K motion sequences spanning
$40$ classes.
HumanAct12~\cite{guo2020action2motion} contains $1,191$ motions
across $12$ categories.

We further introduce a novel setting: zero-shot action-to-motion, where the 
action labels are completely unseen during training.
NTU94~\cite{shahroudy2016ntu,liu2019ntu} is used for this purpose.
Specifically, it is a 94-class single-person subset of the NTU RGB+D 120
dataset~\cite{shahroudy2016ntu,liu2019ntu}, excluding the $26$ categories of
multi-person actions.
The NTU94 dataset consists of $89$K sequences of human motions, and we randomly
choose $63$ classes as the seen classes for training, and use the remaining $31$
classes for the zero-shot action-to-motion synthesis evaluation.

\noindent \textbf{Evaluation Metrics} \quad
For the text-to-motion evaluation, five metrics are employed, as outlined
in~\cite{guo2022generating}: \emph{Fr\'echet Inception Distance (FID)}
quantifies the disparity between generated and ground truth motion distributions
in the latent space; \emph{Diversity} evaluates the variation in the generated
motion distribution; \emph{Multimodality} measures the average variance given a
single text prompt; \emph{R-Precision} and \emph{Multimodal-Dist} assess the
relevance of generated motion to the textual prompt.
For the R-Precision, we use the top 3 by default.
As our primary goal is to improve the motion synthesis quality, we use FID as
the prior metric.

For the action-to-motion evaluation, we use four metrics: FID, Diversity,
Multimodality and classification accuracy from a pretrained motion classifier.

\begin{table*}[h!]
\centering
\caption{Quantitative results comparison on the KIT test set.}
\label{tab:kit}
\resizebox{0.675\linewidth}{!}{%
\begin{tabular}{c|ccccc}
    \hline 
    Method & FID $\downarrow$ & Diversity $\rightarrow$ & MultiModality $\uparrow$ & R Precision (top3) $\uparrow$ & MultiModal Dist $\downarrow$\tabularnewline
    \hline 
    Real Motion & $0.031^{\pm .004}$ & $11.08^{\pm .097}$ & - & $0.779^{\pm .006}$ & $2.788^{\pm .012}$\tabularnewline
    \hline 
    Language2Pose~\cite{ahuja2019language2pose} & $6.545^{\pm .072}$ & $9.073^{\pm .100}$ & - & $0.483^{\pm .005}$ & $5.147^{\pm .030}$\tabularnewline
    Text2Gesture~\cite{bhattacharya2021text2gestures} & $12.12^{\pm .183}$ & $9.334^{\pm .079}$ & - & $0.338^{\pm .005}$ & $6.964^{\pm .029}$\tabularnewline
    Hier~\cite{ghosh2021synthesis} & $5.203^{\pm .107}$ & $9.563^{\pm .072}$ & - & $0.531^{\pm .007}$ & $4.986^{\pm .027}$\tabularnewline
    T2M~\cite{guo2022generating} & $2.770^{\pm .109}$ & $10.91^{\pm .119}$ & $1.482^{\pm .065}$ & $\bm{0.693^{\pm .007}}$ & $\bm{3.401^{\pm .008}}$\tabularnewline
    MoCoGAN~\cite{tulyakov2018mocogan} & $82.69^{\pm .242}$ & $3.092^{\pm .043}$ & $0.250^{\pm .009}$ & $0.063^{\pm .003}$ & $10.47^{\pm .012}$\tabularnewline
    Dance2Music~\cite{lee2019dancing} & $115.4^{\pm .240}$ & $0.241^{\pm .004}$ & $0.062^{\pm .002}$ & $0.086^{\pm .003}$ & $10.40^{\pm .016}$\tabularnewline
    Ours & \textbf{$\bm{0.472^{\pm .029}}$} & $\bm{10.957^{\pm .092}}$ & $\bm{2.049^{\pm .086}}$ & $0.390^{\pm .006}$ & $9.161^{\pm .027}$\tabularnewline
    \hline 
\end{tabular}%
}
\end{table*}

\begin{table*}[h]
\centering
\caption{Quantitative results comparison on the UESTC dataset.}
\label{tab:uestc}
\resizebox{0.6\linewidth}{!}{%
\begin{tabular}{c|ccccc}
    \hline 
    Method & FID (train) $\downarrow$ & FID (test) $\downarrow$ & Accuracy $\uparrow$  & Diversity $\rightarrow$ & MultiModality $\rightarrow$\tabularnewline
    \hline 
    Real Motion & $2.92^{\pm .26}$ & $2.79^{\pm .29}$ & $0.988^{\pm .01}$ & $33.44^{\pm .320}$ & $14.16^{\pm .06}$\tabularnewline
    \hline 
    Action2Motion~\cite{guo2020action2motion} & $21.02^{\pm 2.51}$ & $24.08^{\pm 2.17}$ & $0.889^{\pm .01}$ & $30.47^{\pm .33}$ & $13.46^{\pm .03}$\tabularnewline
    ACTOR~\cite{petrovich2021action} & $20.49^{\pm 2.31}$ & $23.43^{\pm 2.20}$ & $0.911^{\pm .00}$ & $31.96^{\pm .36}$ & $\bm{14.66^{\pm .03}}$\tabularnewline
    INR~\cite{cervantes2022implicit} & $9.55^{\pm .06}$ & $15.00^{\pm .09}$ & $\bm{0.941^{\pm .00}}$ & $31.59^{\pm .19}$ & $14.68^{\pm .07}$\tabularnewline
    Ours & $\bm{6.68^{\pm .04}}$ & $\bm{9.67^{\pm .17}}$ & $0.934^{\pm .01}$ & $\bm{32.22^{\pm .13}}$ & $15.43^{\pm .06}$\tabularnewline
    \hline 
\end{tabular}%
}
\end{table*}

\begin{table*}[h]
\begin{minipage}[h]{0.54\linewidth}
\centering
\caption{Quantitative results comparison on the HumanAct12 dataset.}
\label{tab:humanact12}
\resizebox{\linewidth}{!}{%
\begin{tabular}{c|cccc}
    \hline 
    Method & FID (train) $\downarrow$ & Accuracy $\uparrow$ & Diversity $\rightarrow$ & MultiModality $\rightarrow$\tabularnewline
    \hline 
    Real Motion & $0.09^{\pm .01}$ & $0.997^{\pm .10}$ & $6.85^{\pm .05}$ & $2.45^{\pm .04}$\tabularnewline
    \hline 
    Action2Motion~\cite{guo2020action2motion} & $2.45^{\pm .08}$ & $0.923^{\pm .02}$ & $7.03^{\pm .04}$ & $2.87^{\pm .04}$\tabularnewline
    ACTOR~\cite{petrovich2021action} & $0.12^{\pm .00}$ & $0.955^{\pm .08}$ & $\bm{6.84^{\pm .03}}$ & $2.53^{\pm .02}$\tabularnewline
    INR~\cite{cervantes2022implicit} & $0.09^{\pm .00}$ & $0.973^{\pm .00}$ & $6.88^{\pm .05}$ & $2.57^{\pm .04}$\tabularnewline
    Ours & $\bm{0.09^{\pm .01}}$ & $\bm{0.976^{\pm .01}}$ & $6.82^{\pm .02}$ & $\bm{2.52^{\pm .03}}$\tabularnewline
    \hline 
\end{tabular}%
}
\end{minipage}
\hfill
\begin{minipage}[h]{0.44\linewidth}
\centering
\caption{Quantitative results comparison on the NTU94 dataset. Closed-set (seen
classes) and zero-shot (unseen classes) synthesis results are respectively
reported.}
\label{tab:ntu-zero-shot}
\vspace{-0.3cm}
\resizebox{\linewidth}{!}{%
\begin{tabular}{c|cc|cc}
    \hline 
    \multirow{2}{*}{Method} & \multicolumn{2}{c|}{Seen Classes} & \multicolumn{2}{c}{Unseen Classes}\tabularnewline
    \cline{2-5} \cline{3-5} \cline{4-5} \cline{5-5} 
     & FID (train) $\downarrow$ & Acc.$\uparrow$ & FID (train) $\downarrow$ & Acc.$\uparrow$\tabularnewline
    \hline 
    Random Generator & $315.55^{\pm0.16}$ & $0.016^{\pm.00}$ & $319.53^{\pm0.92}$ & $0.032^{\pm.00}$\tabularnewline
    Action2Motion~\cite{guo2020action2motion} & $138.62^{\pm0.02}$ & $0.817^{\pm.00}$ & - & -\tabularnewline
    ACTOR~\cite{petrovich2021action} & $136.72^{\pm0.03}$ & $0.823^{\pm.00}$ & - & -\tabularnewline
    Action2Motion~\cite{guo2020action2motion} w/ CLIP & $131.48^{\pm0.01}$ & $0.833^{\pm.00}$ & $221.57^{\pm0.21}$ & $0.108^{\pm.00}$\tabularnewline
    ACTOR~\cite{petrovich2021action} w/ CLIP & $128.80^{\pm0.02}$ & $0.835^{\pm.00}$ & $205.80^{\pm0.24}$ & $0.116^{\pm.00}$\tabularnewline
    Ours & \textbf{$\bm{42.30^{\pm0.01}}$} & \textbf{$\bm{0.869^{\pm.00}}$} & \textbf{$\bm{112.95^{\pm0.13}}$} & \textbf{$\bm{0.153^{\pm.00}}$}\tabularnewline
    \hline 
\end{tabular}%
}
\end{minipage}
\end{table*}

\noindent \textbf{Implementation Details} \quad
Our \abbrmodelname{} is trained with the AdamW optimizer for $50$K epochs, with
an initial learning rate of $10^{-4}$ decay at the $40$K-th step by a factor of
$10$.
All of the hyperparameters are determined via a grid search:
$w_{\text{KL}}=w_{\text{div}}=w_{\text{spa}}=10^{-2}$.
We set the hidden dimension $D=512$, the number of Transformer encoder layers
and decoder layers to $8$, and the number of attention head $h=8$.
We set the number of atomic actions $N=256$ for the action-to-motion task, and
$N=1024$ for the text-to-motion task.
We follow existing methods~\cite{tevet2022motionclip} to use
CLIP~\cite{radford2021learning,tevet2022human} for the language embedding
extraction.
For the text-to-motion task, we generate $120$ frames, and for the
aciton-to-motion task we generate $60$ frames.
We follow existing
methods~\cite{guo2020action2motion,petrovich2021action,guo2022generating} to
select the classifier for metric computation for each task.
For the zero-shot action-to-motion task, we use a pretrained ST-GCN
classifier~\cite{yan2018spatial} to compute the metrics.

\subsection{Text-to-motion Evaluation}

\noindent \textbf{HumanML3D} \quad
We compare our \abbrmodelname{} with existing methods on the HumanML3D dataset
in~\cref{tab:humanml3d}.
Our approach achieves strong FID, signifying a closer match between
generated and ground truth motion distributions.
Moreover, our \abbrmodelname{} displays superior Diversity and MultiModality
scores, showcasing its ability to generate a wide range of motion sequences.
Our method is also competitive in terms of R Precision and MultiModal Dist,
indicating good alignment between motion and language.
Collectively, these results emphasize the strengths of \abbrmodelname{} in
generating high-quality and diverse human motion sequences, surpassing previous
methods across various aspects.

\noindent \textbf{KIT} \quad
The results on the KID dataset further emphasize the superiority of our
\abbrmodelname{} in motion synthesis.
As illustrated in~\cref{tab:kit}, our method significantly outperforms previous
approaches in terms of FID, Diversity, and MultiModality, indicating a
substantial reduction in the discrepancy between generated and ground truth
motion distributions and the ability to produce diverse motions.
Although our \abbrmodelname{} demonstrates promising synthesis outcomes, there
is still room for improvement in aligning textual descriptions with the
generated motion, which could be addressed by incorporating more advanced
language models for language embedding.
Overall, these results highlight \abbrmodelname{}'s effectiveness in generating
high-quality and diverse human motion sequences.

\subsection{Action-to-motion Evaluation}

\noindent \textbf{UESTC} \quad
Our experimental results on the UESTC dataset provide a compelling demonstration
of \abbrmodelname{}'s effectiveness in the action-to-motion synthesis task, as
shown in~\cref{tab:uestc}.
Among the four metrics, FID is the most important indicator in evaluating the
overall synthesis quality.
Our \abbrmodelname{} achieves the lowest FID on the training and testing
subsets, significantly outperforming the best results from previous methods.
This demonstrates that our method generates more realistic motion sequences
compared to the competing methods.
In terms of accuracy, \abbrmodelname{} attains a strong accuracy of $0.934$,
indicating that our method is accurate in generating motion sequences that
correspond to the given action label.
Regarding the diversity of the generated motions, \abbrmodelname{} outperforms
previous methods in diversity and achieves competitive multi-modality score.
These results illustrate our method's capability to generate realistic, diverse
and rich motion sequences.

\noindent \textbf{HumanAct12} \quad
The experimental results on the HumanAct12 dataset in~\cref{tab:humanact12}
further validate the effectiveness of our \abbrmodelname{} in the
action-to-motion synthesis task.
Our method and INR~\cite{cervantes2022implicit} achieve the same lowest FID of $0.09$, closely approximating the real motion score
of $0.09$.
This consistency between the UESTC and HumanAct12 datasets highlights the
robustness of our method in generating realistic motion sequences.
Our \abbrmodelname{} also achieves the highest classification accuracy, showing
its ability to generate motions that correspond to the action labels across
different datasets.
We also achieve promising diversity and multi-modality scores.
Overall, the consistent and superior performance of \abbrmodelname{} across both
UESTC and HumanAct12 datasets emphasizes its effectiveness and robustness in the
action-to-motion synthesis task.

\begin{figure}[t]
    \centering
    \subfloat[Seen classes.]{
    \includegraphics[width=\linewidth]{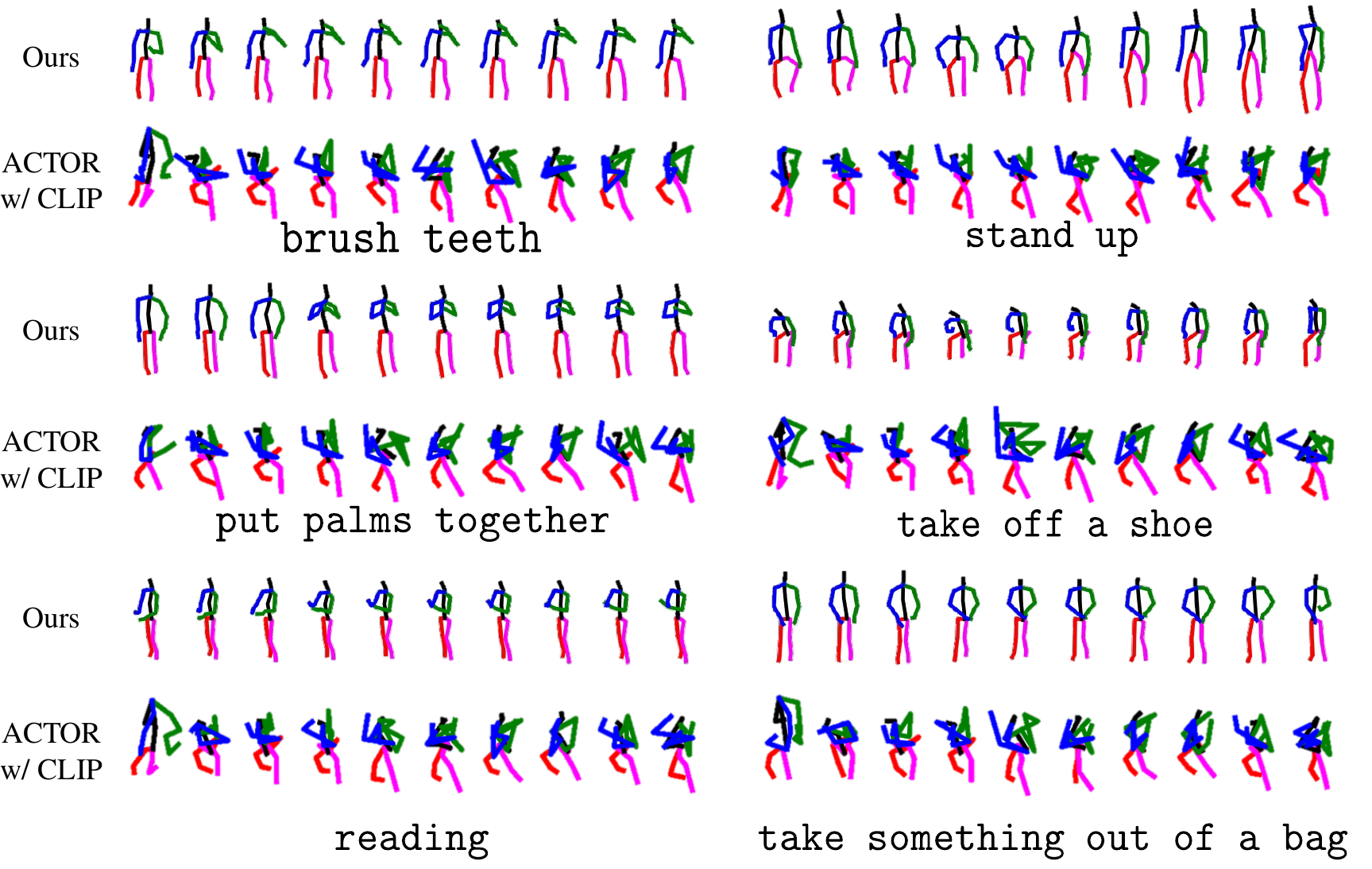}
    }

    \subfloat[Unseen classes.]{
    \includegraphics[width=\linewidth]{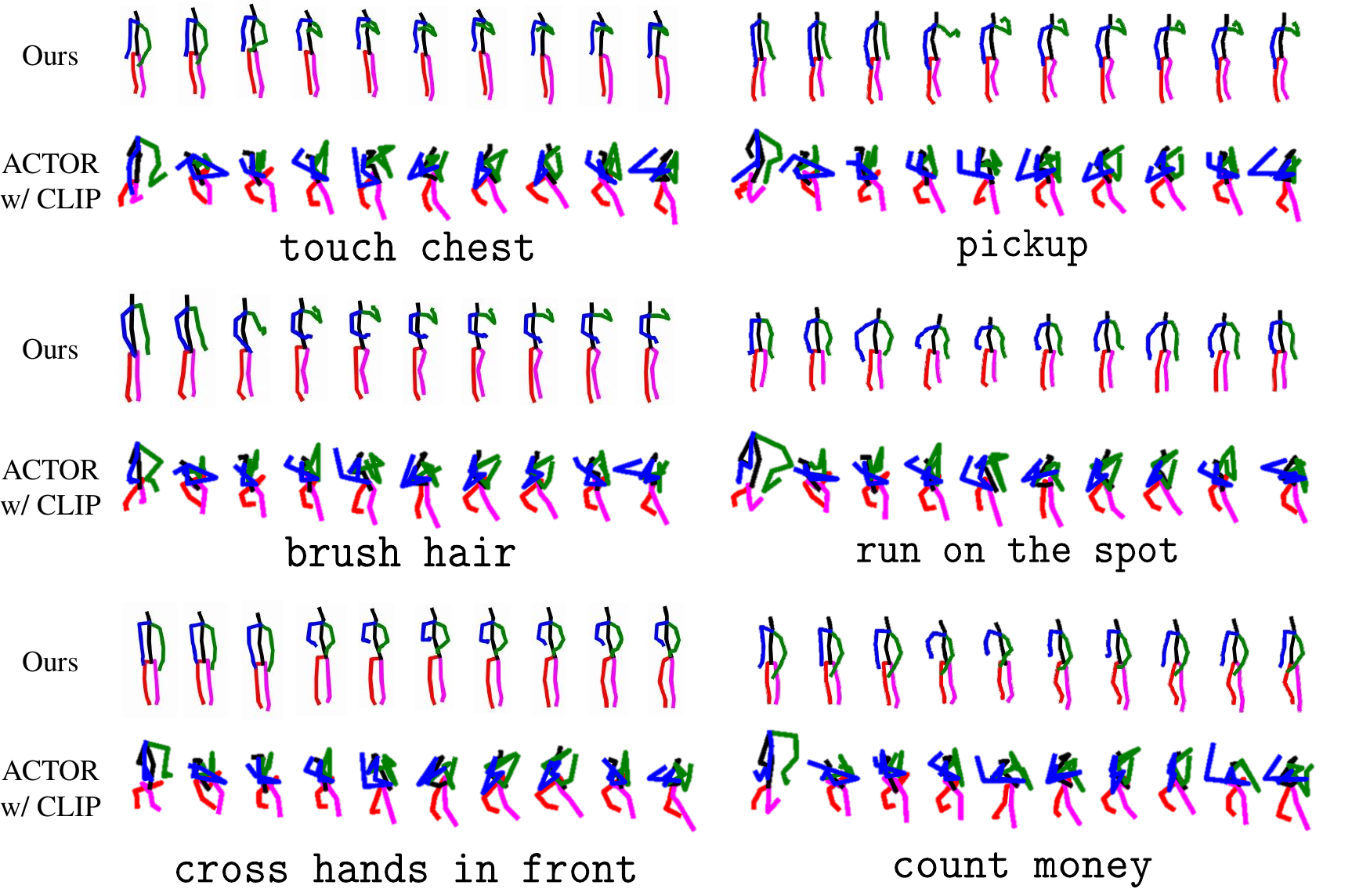}
    }
    \caption{Zero-shot action-to-motion qualitative results comparison with
    ACTOR~\cite{petrovich2021action} on the NTU94 dataset.}
    \label{fig:ntu}
\end{figure}

\subsection{Zero-shot Action-to-motion Synthesis}
Apart from the common text-to-motion and action-to-motion settings, we introduce
a novel setting: zero-shot action-to-motion synthesis, where the action labels
for inference are entirely unseen during training.
This new setting aims to evaluate the model's ability to generalize to unseen
actions, crucial for practical applications with novel action descriptions.
Assessing performance in the zero-shot action-to-motion setting enables a better
understanding of the model's potential to synthesize realistic and diverse human
motion sequences in real-world scenarios.

We evaluate our zero-shot action-to-motion synthesis performance on the NTU94
dataset, and the results are listed in~\Cref{tab:ntu-zero-shot}.
Two previous methods are included for comparison, \ie{},
ACTOR~\cite{petrovich2021action} and Action2Motion~\cite{guo2020action2motion}.
Note they take as input one-hot label vectors to output motion sequences of
certain classes, thus they cannot synthesize actions of unseen classes (denoted
as ``$-$'' in~\Cref{tab:ntu-zero-shot}).
To achieve zero-shot synthesis, we replace the one-hot label vector input in
with CLIP text embedding, denoted as ``w/ CLIP''.
We make the following observations from the results:
(1)~Our \abbrmodelname{} significantly outperforms previous methods in both seen
class and unseen class synthesis tasks.
Surprisingly, our FID on unseen classes is even lower than the ACTOR's FID on
seen classes, which demonstrates the effectiveness of our method.
(2)~Our method exhibits lower variance compared with previous methods,
indicating a more stable synthesis results.
(3)~Though previous methods show promising results on small-scale human motion
datasets (\eg{}, NTU13 and HumanAct12), they show poor capability in large-scale
datasets such as NTU94.

We further compare the qualitative results with ACTOR~\cite{petrovich2021action}
w/ CLIP in~\cref{fig:ntu}.
First, our \abbrmodelname{} can successfully generate seen classes, while ACTOR
shows poor qualitative results, coinciding previous
findings~\cite{song2022actformer} that several previous methods only work well
on simple human motion datasets consisting of $\sim 10$ classes.
The high FID and high accuracy of previous
methods~\cite{guo2020action2motion,petrovich2021action} show that they focus on
learning trivial motions for classification, instead of human-recognizable
motions. 
Second, we observe our method generates realistic motions on unseen classes when
there are certain body parts in the textual description, \eg{}, ``\texttt{touch
\underline{chest}}'', ``\texttt{brush \underline{hair}}'', or ``\texttt{cross
\underline{hands} in front}''.
In contrast, ACTOR cannot generate meaningful human motions in such cases.

\begin{figure}[t]
    \centering
    \includegraphics[width=\linewidth]{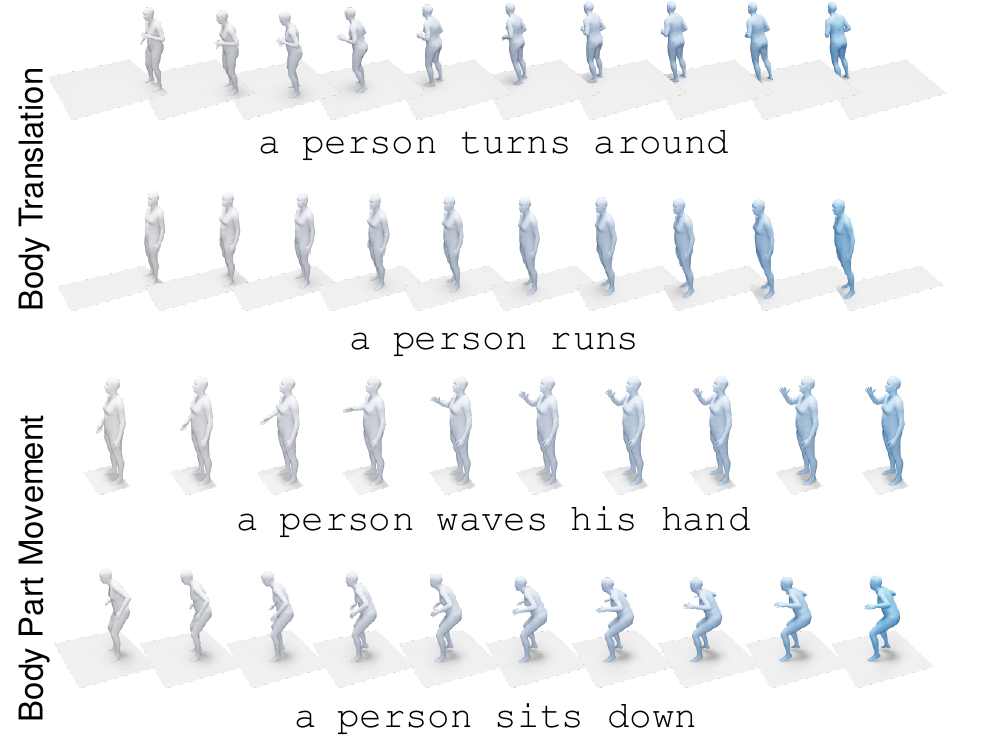}
    \caption{Visualization of learned atomic actions on the HumanML3D dataset.
    For each textual prompt, we visualize the atomic action that has the highest
    attention value to it.}
    \label{fig:atomic-action}
\end{figure}

\begin{table}[t]
\centering
\caption{Ablation study on the atomic action codebook on the KIT
dataset. The variant of model w/o codebook is realized by implementing the
decoder with a Transformer encoder, where the query, key and value are the
same.}
\label{tab:atomic-action}
\resizebox*{\linewidth}{!}{%
\begin{tabular}{ccc|ccc}
    \hline 
    \#Atom & $\mathcal{{L}_{\text{{div}}}}$ & $\mathcal{{L}_{\text{{spa}}}}$ & FID $\downarrow$ & Diversity $\rightarrow$ & R Precision $\uparrow$\tabularnewline
    \hline 
    \multicolumn{3}{c|}{CVAE baseline} & $6.87^{\pm.021}$ & $9.102^{\pm.072}$ & $0.280^{\pm.024}$\tabularnewline
    \hline 
    256 & - & - & $4.57^{\pm.031}$ & $9.204^{\pm.038}$ & $0.295^{\pm.039}$\tabularnewline
    512 & - & - & $2.42^{\pm.023}$ & $9.673^{\pm.057}$ & $0.303^{\pm.034}$\tabularnewline
    1024 & - & - & $1.43^{\pm.038}$ & $10.327^{\pm.052}$ & $0.337^{\pm.014}$\tabularnewline
    2048 & - & - & $1.47^{\pm.033}$ & $10.311^{\pm.062}$ & $0.356^{\pm.023}$\tabularnewline
    \hline 
    1024 & $\checkmark$ & - & $0.873^{\pm.021}$ & $10.937^{\pm.069}$ & $0.381^{\pm.003}$\tabularnewline
    1024 & - & $\checkmark$ & $0.732^{\pm.039}$ & $10.688^{\pm.073}$ & $0.384^{\pm.012}$\tabularnewline
    1024 & $\checkmark$ & $\checkmark$ & \textbf{$\bm{0.472^{\pm .029}}$} & $\bm{10.957^{\pm .092}}$ & $0.390^{\pm .006}$\tabularnewline
    \hline 
\end{tabular}%
}
\end{table}

\begin{table}[t]
\centering
\caption{Ablation study masked motion modeling curriculum learning on the KIT dataset.}
\label{tab:curriculum-learning}
\resizebox*{\linewidth}{!}{%
\begin{tabular}{cc|ccc}
    \hline 
    Mask ratio $r$ & Learning Scheme & FID $\downarrow$ & Diversity $\rightarrow$ & R Precision $\uparrow$\tabularnewline
    \hline 
    \multicolumn{2}{c|}{w/o masked modeling} & $2.48^{\pm.031}$ & $10.342^{\pm.031}$ & $0.345^{\pm.021}$\tabularnewline
    \hline 
    25\% & \multirow{3}{*}{$g(l)=r$} & $1.22^{\pm.030}$ & $10.473^{\pm.037}$ & $0.344^{\pm.008}$\tabularnewline
    50\% &  & $0.76^{\pm.049}$ & $10.659^{\pm.051}$ & $0.372^{\pm.009}$\tabularnewline
    75\% &  & $1.48^{\pm.033}$ & $10.551^{\pm.034}$ & $0.388^{\pm.013}$\tabularnewline
    \hline 
    \multirow{3}{*}{50\%} & $g(l)=rl/L$ & \textbf{$\bm{0.472^{\pm .029}}$} & $\bm{10.957^{\pm .092}}$ & $\bm{0.390^{\pm .006}}$\tabularnewline
     & $g(l)=r(l/L)^{2}$ & $0.621^{\pm.031}$ & $10.683^{\pm.041}$ & $0.373^{\pm.009}$\tabularnewline
     & $g(l)=re^{l/L-1}$ & $0.583^{\pm.024}$ & $10.590^{\pm.057}$ & $0.387^{\pm.007}$\tabularnewline
    \hline 
\end{tabular}%
}
\end{table}

\subsection{Ablation Study}

\noindent \textbf{Atomic Action} \quad
First, we quantitatively analyze the effect of atomic actions
in~\cref{tab:atomic-action}.
As the number of atomic actions increases, the FID decreases, and both the
Diversity and R Precision improve, indicating better motion generation quality.
However, further increasing the number of atomic actions to $2048$ does not
yield substantial improvements.
The combination of diversity and sparsity constraints leads to the best
performances, suggesting that they are essential components for generating
high-quality, diverse human motion sequences in our model.

We further visually analyze the learned atomic actions
in~\cref{fig:atomic-action}.
Upon examination, we find that these atomic actions primarily represent two
types of motion: body translation and specific body part movements.
Body translation atomic actions typically involve motions like running, or
actions involve directional changes, where the entire body is engaged in
coordinated movement, see the first two examples of~\cref{fig:atomic-action}.
On the other hand, specific body part movements focus on the motion of a
particular body part, such as waving a hand, or sitting down, see the last two
examples of~\cref{fig:atomic-action}. 
This distinction highlights the versatility of the learned atomic actions, as
they capture both global and local motion patterns.
As a result, our approach can effectively synthesize complex human motions by
combining these diverse atomic actions in a meaningful and contextually
appropriate manner.

\noindent \textbf{Masked Motion Modeling Curriculum Learning} \quad
The results of the masked motion modeling curriculum learning are illustrated
in~\cref{tab:curriculum-learning}.
When compared to the baseline without masked modeling, incorporating masked
modeling consistently leads to a reduction in FID and an improvement in
Diversity and R Precision, and achieves the best results at a mask ratio of
$50\%$, indicating the benefit of this approach for motion generation quality.
Moreover, the learning scheme that follows a linear progression $g(l)=rl/L$
demonstrates the best performance across all metrics, suggesting that a gradual
exposure to the complexity of motion reconstruction during training is an
effective strategy for enhancing the model's synthesis capabilities.

\begin{figure}[t]
    \includegraphics[width=\linewidth]{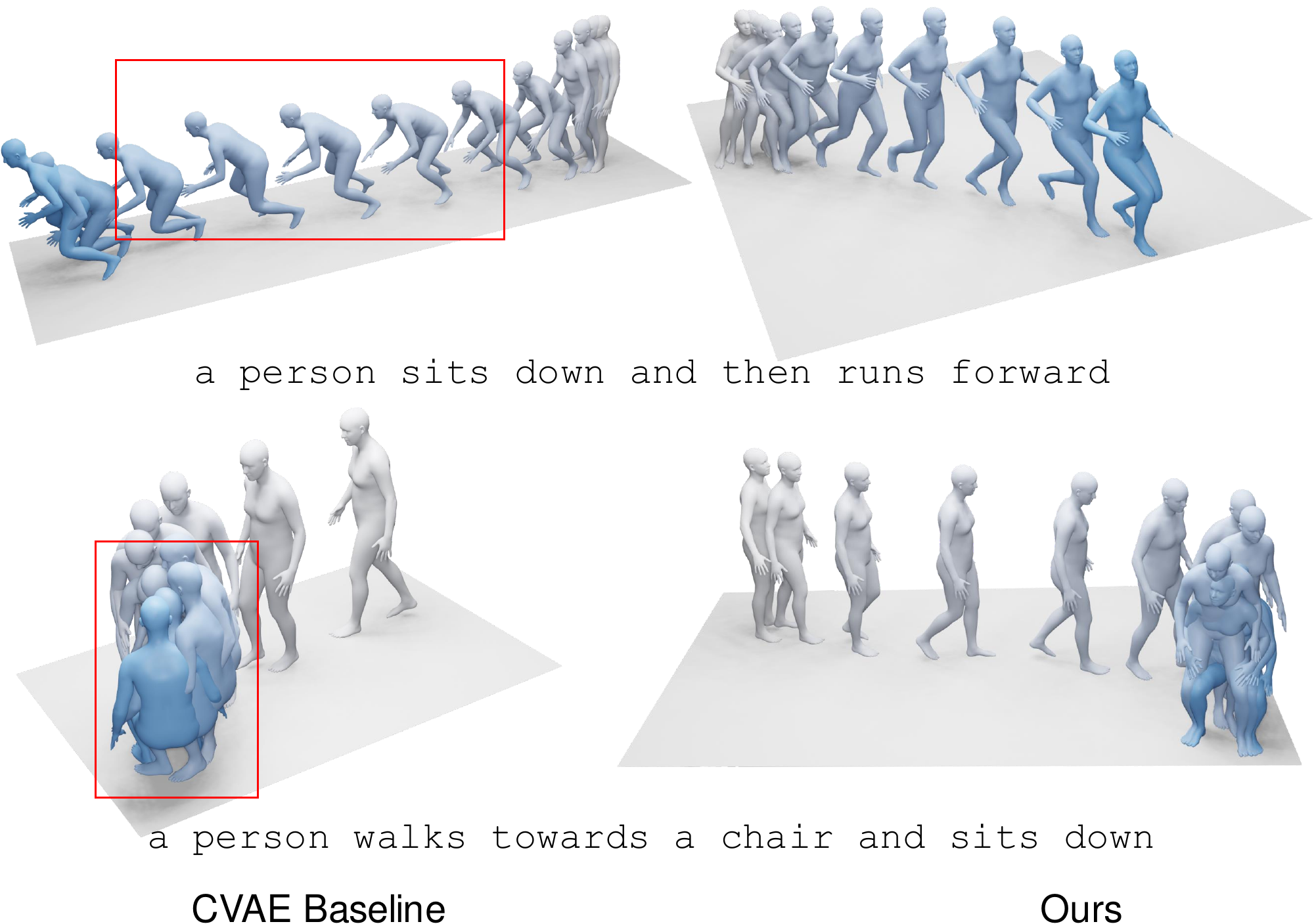}
    \caption{Qualitative comparison between our CVAE baseline and our
    \abbrmodelname{}.}
    \label{fig:ablation-qua}
\end{figure}

\noindent \textbf{Qualitative Results} \quad
We showcase qualitative results in~\cref{fig:ablation-qua}, highlighting the
differences between our \abbrmodelname{} and the CVAE baseline.
In the first example, the human figure transitioning between "\texttt{sitting}"
and "\texttt{running}" appears tilted in the CVAE result.
For the second CVAE example, the person continues to rotate even after sitting
down.
In contrast, our final \abbrmodelname{} generates a more natural and realistic
motion sequences, demonstrating the effectiveness of our proposed atomic action
and curriculum learning.

\section{Conclusion}
\label{sec:conclusion}

We present \abbrmodelname{}, a novel approach for language-guided human motion
synthesis that addresses the limitations of previous methods by leveraging
atomic action decomposition and a curriculum learning strategy.
\abbrmodelname{}'s transformer-based CVAE framework effectively decomposes
complex actions into atomic components, allowing for the generation of diverse
and coherent motion sequences.
The incorporation of curriculum learning further enhances the model's ability to
capture dependencies in motion sequences, resulting in smooth and natural motion
transitions between different actions.
Our comprehensive evaluation demonstrates \abbrmodelname{}'s superior
performance over existing approaches in the text-to-motion and action-to-motion
synthesis tasks.

\begin{acks}
This work is supported in part by the Defense Advanced Research Projects Agency (DARPA) under Contract No.~HR001120C0124, the National Science Foundation (NSF) IIS-2008532, NSF Award \#1846076, and the Institute of Education Sciences (IES), U.S. Department of Education (ED) through Award \#2229873. Any opinions, findings and conclusions or recommendations expressed in this material are those of the author(s) and do not necessarily reflect the views of the DARPA, the NFS, the IES, or the ED.
\end{acks}

\vfill\eject
\bibliographystyle{ACM-Reference-Format}
\bibliography{ref}

\end{document}